\documentclass[journal]{IEEEtran}
\usepackage[numbers]{natbib}
\usepackage{bm}
\usepackage{subfigure}
\usepackage{amsmath}
\usepackage{amsfonts}
\usepackage{graphicx}
\usepackage{booktabs}
\usepackage{multirow}
\usepackage{longtable}
\usepackage{algorithm}

\usepackage{algorithmic}

\renewcommand{\baselinestretch}{.89}

\ifCLASSINFOpdf
\else
\fi
\begin{document}
%
\title{Hybrid Reinforced Medical Report Generation with M-Linear Attention and Repetition Penalty}
%
%
%


  \author{Wenting Xu, Zhenghua Xu, Junyang Chen, Chang Qi, Thomas Lukasiewicz
\thanks{This work was supported by the National Natural Science Foundation of China under the grant 61906063, by the Natural Science Foundation of Hebei Province, China, under the grant F2021202064, by the Natural Science Foundation of Tianjin City, China, under the grant 19JCQNJC00400, by the “100 Talents Plan” of Hebei Province, China, under the grant E2019050017, and by the Yuanguang Scholar Fund of Hebei University of Technology, China. This work was also partially supported by the AXA Research Fund.}
\thanks{Wenting Xu, Zhenghua Xu and Chang Qi are with the State Key Laboratory of Reliability and Intelligence of Electrical Equipment and Tianjin Key Laboratory of Bioelectromagnetic Technology and Intelligent Health, Hebei University of Technology, China. (Corresponding author: Zhenghua Xu, e-mail: zhenghua.xu@hebut.edu.cn.)}
\thanks{Junyang Chen is with the College of Computer Science and Software Engineering, Shenzhen University, China}
\thanks{Thomas Lukasiewicz is with the Department of Computer Science, University of Oxford, United Kingdom.}
}

%
%

\markboth{IEEE TRANSACTIONS ON NEURAL NETWORKS AND LEARNING SYSTEMS,~Vol.~xx, No.~x, xxx~2022}%
{Xu \MakeLowercase{\textit{et al.}}: Bare Demo of IEEEtran.cls for IEEE Journals}
%



\maketitle

\begin{abstract}
To reduce doctors' workload, deep-learning-based automatic
medical report generation has recently attracted more and
more research efforts,
where deep convolutional neural networks (CNNs) are employed to encode the input images, and recurrent neural networks (RNNs) are used to decode the visual features into medical reports automatically.  
However, these state-of-the-art methods mainly suffer from three shortco\-mings: (i)~incomprehensive optimization,  
(ii) low-order and unidimensional attention mechanisms,   
and (iii) repeated generation.  
In this article, we propose a hybrid reinforced medical report generation method with m-linear attention and repetition penalty mechanism (HReMRG-MR) to overcome these problems. Specifically, a hybrid reward with different weights is employed to remedy the limitations of single-metric-based rewards. We also propose a search algorithm with linear complexity to approximate the best weight combination. Furthermore, we use 
m-linear attention modules  to explore high-order feature interactions and to achieve multi-modal reasoning,
while a repetition penalty applies penalties to repeated
terms during the model’s training process. Extensive experimental studies  on two public datasets show that HReMRG-MR greatly outperforms
the state-of-the-art baselines in terms of all metrics. We also conducted a series of ablation experiments to prove the effectiveness of all our proposed components. We also performed a~reward search toy experiment  to give evidence that our proposed search approach can  significantly reduce the search time while approximating the best performance. 
\end{abstract}


%
\IEEEpeerreviewmaketitle

\vspace{-0.5em}
\section{Introduction}
%
%
%
%

\IEEEPARstart{I}{n} recent years, medical imaging has become the most commonly used medical examination method in disease diagnosis.  It produces reports that are paragraph-based documents issued by radiologists after examinations. These reports describe the important medical findings observed on the corresponding medical images and emphasize the abnormalities, along with the sizes and locations of detected lesions. 
However, due to the increasing number of patients and the shortage of experienced radiologists, a radiologist may have to conduct dozens or sometimes even hundreds of medical imaging examinations and then write the same number of reports every day, which makes the radiologists overloaded and may lead to increasing misdiagnosis.
Therefore, it is a compelling demand to find promising methods to generate medical reports automatically.

Existing deep-learning-based medical report generation methods mainly adopt the encoder-decoder architecture~\cite{Wang_2018,jing-etal-2018-automatic,xue2018multimodal}, where deep convolutional neural networks (CNNs)  encode the input medical images, and then recurrent neural networks (RNNs), e.g., long short-term memory (LSTM),  as  decoder  generate medical reports automatically.
However, such encoder-decoder models inevitably suffer from the problem of sentence coherence:
Cross-entropy is widely used in these methods for optimization, however, it only focuses on  word-level errors but ignores the inter-word connections;
since the generated reports consist of long sentences, the coherence of their resulting sentences is usually not satisfactory.
Therefore,  several recent works use reinforcement learning (RL)  \cite{10.1007/978-3-030-32692-0_77,liu2019clinically} to enhance the presentation of the generated reports.

Despite achieving some improvements, the existing reinforcement learning methods still suffer from three problems as follows. (i) Incomprehensive optimization goals: Current reinforcement-learning-based approaches tend to adopt one-or-two metrics like CIDEr as the reward to optimize the performance in this aspect. However, as a long-text generation task, its performance cannot be evaluated by only one or two metrics comprehensively. Therefore, simply employing one-or-two metrics can only lead to partial optimization, failing to perform an overall optimization.    
(ii) Low-order and unidimensional attention mechanisms:  
To raise the model performance, attention is often applied to attach different weights to the features according to their importance, while current medical report generation works only adopt single low-order visual and/or semantic attention, which cannot exploit  channel-wise information and high-order feature interactions of medical images and texts to obtain a fine-grained visual and semantic information, thus failing to catch high-order and multi-modal information~\cite{jing-etal-2018-automatic,xue2018multimodal}. However, high-order information also plays an important role in our task of medical report generation, while the methods above cannot meet the demand. 
(iii) Repeated generation:
As large amounts of our reports are paragraphed normal findings, most works inevitably suffer from the problem of repeated phrases, which thus weakens the coherence and readability of the generated medical reports. 


In this paper, to overcome the above problems, we propose a \textbf{H}ybrid \textbf{Re}inforced \textbf{M}edical \textbf{R}eport \textbf{G}eneration method with \textbf{M}-linear attention and \textbf{R}epetition penalty mechanisms (abbreviated by HReMRG-MR). Compared to existing medical report generation, the proposed HReMRG-MR consists of three improvements. 
First, since   CNN-RNN-based methods employing cross-entropy as  loss fail  to consider sentence-level accuracy, leading to partial optimization, some works explore to achieve sentence-level accuracy via reinforcement learning. However, they all adopt one-or-two-metric  rewards like CIDEr, failing to achieve an overall optimization in all metrics.
Intuitively, we believe that reinforcement learning with a hybrid reward will close the gap of incomprehensive training goals, contributing to a boost on all metrics. 
As different metrics play different roles, we are supposed to give them different  weights. However, if we adopt trivial methods to compute the weights, the cost of computation is exponential. Therefore, we provide a search solution to search for an optimal  weight combination for a hybrid reward to further boost the overall performance with linear computation costs.


To remedy the problem of the low-order and unidimensional attention mechanisms that existing medical report generation methods adopt, we also propose an m-linear attention mechanism to strengthen the model's capability in intra- and inter-modal reasoning.
The proposed m-linear attention module is a stack of bi-linear pooling-based attention blocks, aiming to calculate the outer product between two feature vectors and enable channel-wise attention. Bi-linear pooling is a feature fusion pattern, which enables second-order feature interactions across all feature elements, thus making an advantage over commonly adopted feature fusion methods in achieving high-order feature interactions. By introducing bi-linear pooling and the stacked architecture, our attention mechanism is capable of exploring high-order feature interactions, so enhancing~the model performance in capturing prominent abnormalities.~It is used during both encoding and decoding procedures to enhance the model's capability of intra- and inter-modal reasoning, and get a better medical report generation performance. 


Furthermore, to overcome the problem of repeated terms, we propose to integrate a repetition penalty module into the process of text generation.  The proposed repetition penalty is an exponential weighted penalty applied to repeated terms. When repetitions occur, the exponential penalty is attached to the high repeated terms, which is believed to help the model generate more diverse terms and increase the coherence and readability of the generated report by suppressing the production of meaningless repetitive words.

Overall, the contributions of this paper are as follows:
	(i) We identify the main limitations of existing medical report generation methods in incomprehensive optimization goals, low-order and unidimensional attention mechanisms, and repeated generation, and propose a Hybrid Reinforced Medical Report Generation method with M-linear attention and Repetition penalty (HReMRG-MR), for more accurate and readable medical report generation. 
	(ii) In HReMRG-MR, we figure out the weakness of the normally used CIDEr-based reward for reinforcement learning, and first propose a hybrid reward with different weights to make the performance more balanced. We also developed a search algorithm with linear complexity to obtain the most suitable weighted reward, which largely reduces the search time. Then, an m-linear attention module is proposed to remedy the limitations of existing low-order and unidimensional attention mechanisms by using high-order inter- and intra-modal feature interactions. Finally, a repetition penalty module is used to enhance the coherence and readability of the generated reports.
	(iii) We conducted extensive experiments on two publicly available medical image report datasets. The experimental results show that our proposed HReMRG-MR model greatly outperforms the state-of-the-art baselines in terms of all metrics. We have also conducted ablation studies to prove that both the m-linear attention and the repetition penalty mechanism are effective and essential for the model to achieve a superior performance. A weight combination search toy experiment for the best combined reward proves the effectiveness of our proposed search. Furthermore, we also conducted a series of parameter search experiments to achieve the best hyperparameters.  

\vspace{-0.5em}
\section{Related Works}\label{sec2}

With rapid advances in reinforcement learning, automatic medical report generation is 
often based on reinforcement learning to boost the performance  \cite{10.1007/978-3-030-32692-0_77,liu2019clinically,li2018hybrid,jing-etal-2019-show}. Specifically, \cite{li2018hybrid} proposes a hybrid retrieval-based model with reinforcement learning to determine whether to generate a sentence via template retrieval or LSTMs. Subsequently, to optimize the non-differentiable and sequence-based test metrics directly, \cite{10.1007/978-3-030-32692-0_77} and  \cite{liu2019clinically} adopt reinforcement learning to directly optimize the score of CIDEr.  Most of them adopt a self-critical sequence training (SCST) method for reinforcement learning, which is the reinforce algorithm that utilizes the output of its own test-time inference algorithm to normalize the rewards that it experiences to tackle the problem of high variance caused by the simple reinforce
algorithm during training \cite{rennie2017self}. Though SCST has achieved a great success in the area of image captioning, the descriptive capacity is limited for generating repeated phrases when encountering long paragraph generation. To alleviate this, in the domain of natural images, \cite{melas2018training} introduces a penalty mechanism when generating captions to reduce the occurrence of recurring terms. Inspired by this work, we also introduce a repetition penalty into our method, rather than simply applying a human-selected penalty factor to the repeated word, we further consider the robustness of the penalty factor and employ an adaptive penalty.
Compared to existing reinforcement learning methods, our proposed HReMRG-MR has three advantages: (i) it utilizes a hybrid reward to make up for the weakness of the traditionally used CIDEr-based reward and to acquire a more balanced performance by increasing readability. (ii) Additionally, a repetition penalty is introduced to help make the model generate more diverse and coherent long sentences. (iii) M-linear attention blocks are also employed to enhance the model's capability in exploring high-order feature interactions and multi-modal reasoning.

	Recent medical report generation approaches 
	have been trying to explore more interactions between images and sentences via attention mechanisms  \cite{10.1007/978-3-030-26763-6_66,Wang_2018,jing-etal-2018-automatic,xue2018multimodal,Chen_2020}. To attend to visual and semantic domains simultaneously, \cite{jing-etal-2018-automatic} proposes a co-attention mechanism that builds attention distributions separately for visual and semantic domains and ignores multi-modal interactions while repetitions tend to be found in the generated reports for the ignorance of contextual coherence. To guarantee the coherence among sentences, \cite{xue2018multimodal} develops a sentence-level attention mechanism to explore multi-modal interactions, which compute the attention distribution over visual regions according to sentence-level semantic features. 
However, all these methods solely explore first-order feature interactions. 
Therefore, in this work, we propose m-linear attention to explore both intra- and inter-modal high-order feature interactions, and further use reinforcement learning to enhance the model's performance. Similar to our work, \cite{Pan2020} also uses a bi-linear attention mechanism, named x-linear attention, to generate more accurate captions for natural images, which is a cascaded two-dimensional (i.e., spatial and channel) attention module. Our work is highly inspired by this work but has a quite improvement focus on the feature fusion pattern. To compare to x-linear attention, we conduct extensive ablation studies to evaluate the effectiveness of our m-linear attention in achieving better medical report generation results.



\begin{figure*}[!t]
	\vspace{-2em}
	\centering {\includegraphics[width=0.9\textwidth]{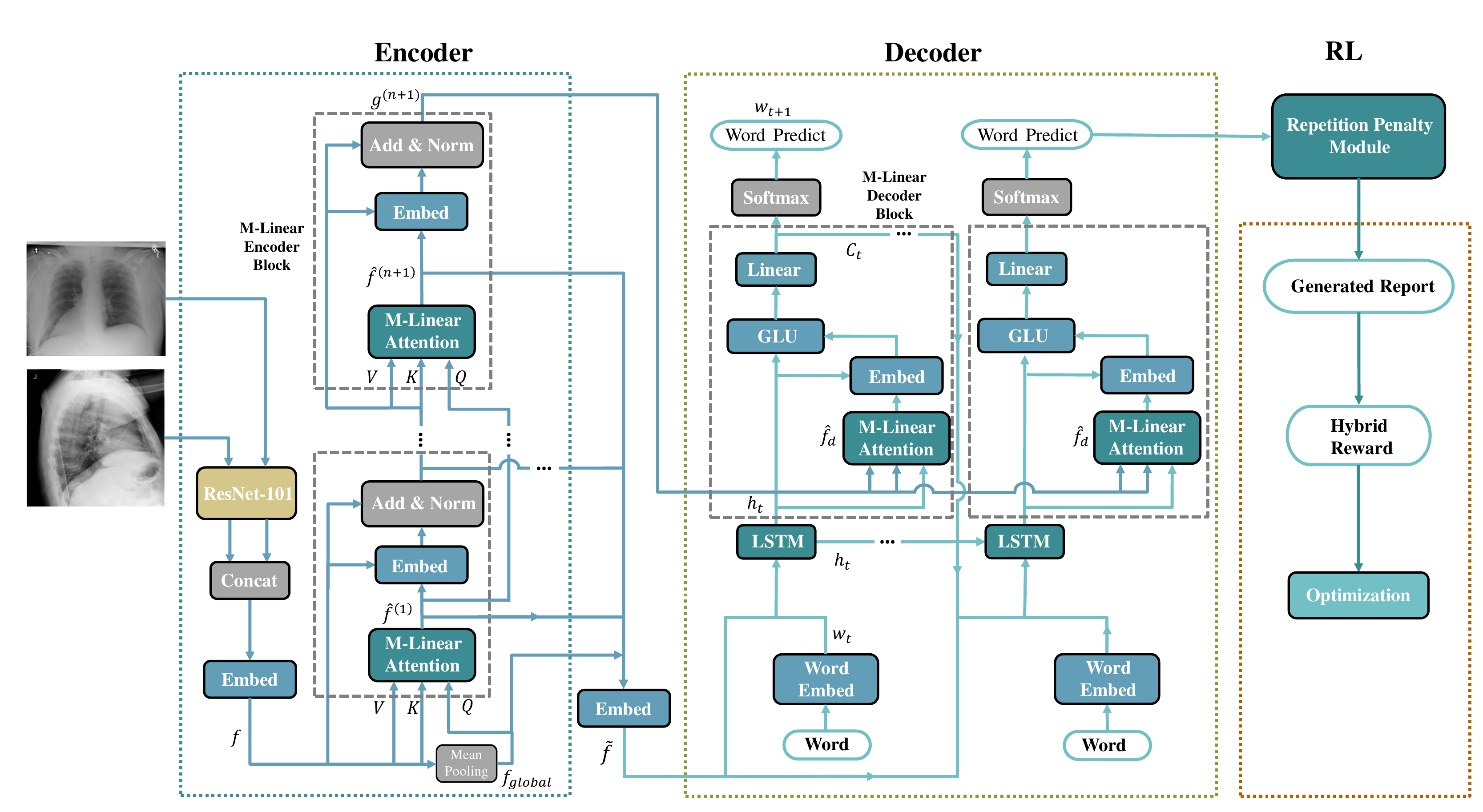}}
	\caption{Architecture of our proposed HReMRG-MR, where Embed denotes a non-linear projection operation, and  GLU denotes gated linear units.
	}\vspace*{-1.5em}
	\label{fig:framework}
\end{figure*}

\vspace{-0.5em}
\section{Methodology}
\label{sec3}

We propose a hybrid reinforced medical report generation method with m-linear attention and repetition penalty (HReMRG-MR). Intuitively, we believe that the use of a hybrid weighted reward will fill the gap of a single CIDEr-based reward, thus generating more readable reports and making the evaluated performance more balanced. 
Besides, we believe that the use of high-order feature interactions will strengthen the model's capacity in single- and multi-modal reasoning, which will enhance the model's performance in terms of the accuracy of the generated reports. 
Moreover, we consider that the combination of the integrated penalty with the repetition term will produce much more diverse sentences, which will also help increase the coherence and readability of the generated diagnosis reports.

Specifically, as shown in Figure~\ref{fig:framework},
images of both frontal and lateral views are encoded through a pretrained ResNet-101 network and concatenated for visual feature extraction. After that, 
to localize prominent abnormalities of chest x-rays and attach the right descriptions to them, 
we embed them and put them into blocks similar to a transformer encoder named m-linear encoder block recursively. Taking the global visual features extracted from medical images $f_{global}$ as input query \textbf{Q} and the regional features $f$ as input value \textbf{V} and key \textbf{K}. A stack of m-linear encoder blocks is  used to calculate the outer product between two feature vectors and enable  channel-wise attention through the squeeze-excitation operation. As such, high-order intra-modal feature interactions are explored during the encoder procedure. 

After that, the embedded attended features from all above layers $\hat{f}$ are combined and sent into LSTMs during the decoder procedure. The output hidden state of the LSTMs together with the final attended visual features $g^(n+1)$ are then sent into an m-linear decoder block to explore multi-modal feature interactions, which is later used for word prediction.

After pretraining with such an encoder-decoder architecture for epochs, reinforcement learning is used to boost the performance, during which we employ the SCST algorithm and use a hybrid weighted reward for optimization. To generate more readable descriptions, we use a repetition penalty module during sentence generation to increase the readability of the generated reports.
We present our network design and implementation details in the following subsections.

%
	

\vspace{-1em}
\subsection{Visual Encoder CNN}
In our work, a ResNet-101-based network pre-trained on ImageNet is employed to extract the global and regional visual features of the different views (frontal and lateral) of a patient's chest x-rays. We resize our input images to 224$\times$224 to keep them consistent with our pre-trained ResNet-based CNN encoder. Then, the regional features $f\in \mathbb{R}^{2048 \times 196}$ (reshaped from 2048$\times$14$\times$14) are extracted from the last convolutional layer of ResNet-101, which represents 196 sub-regions. A global average pooling is applied to the extracted regional features to obtain the global features $f_{global}$. After that, both global and regional features are concatenated according to different views before feeding into the next layers. In this paper, we choose to generate the \textit{Impression} and the \textit{Findings} section. To learn high-order feature interactions both spatially and channel-wise, we introduce an m-linear-attention-based encoder, and detailed information is presented.

\subsubsection{M-Linear Attention}

\begin{figure}[!t]
	\vspace{-1.5em}
	\centering {\includegraphics[width=0.475\textwidth]{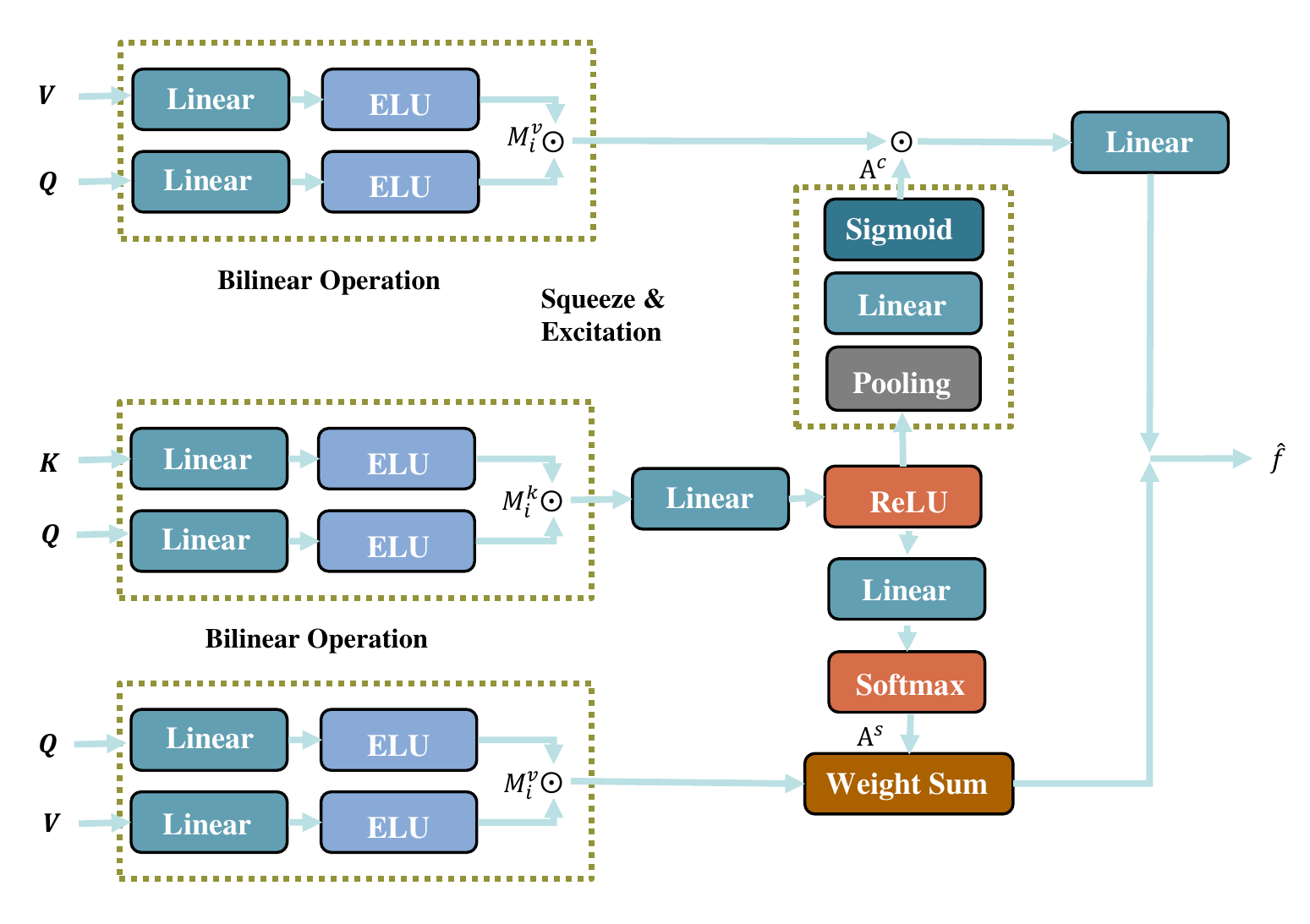}}
	\vspace{-1em}
	\caption{M-linear attention block.}
	\label{fig.attention}\vspace*{-3em}
\end{figure}

Though works in medical report generation have usually focused on improving the attention mechanisms, there are no works exploring the high-order interactions of features. The existing adopted attention mechanisms tend to focus on low-order spatial or semantic attention, ignoring the high-order feature reasoning, thus resulting in the weakness of locating the prominent abnormalities.
To make up for this deficiency, we introduce bi-linear pooling to our attention mechanism.
Bi-linear pooling is first used for fine-grained image classification. It has recently been applied to multi-modal feature fusion in visual question answering, aiming at exploring multi-modal interactions. We use it in our medical report generation to explore high-order intra-modal interactions in medical image features and inter-modal interactions between the radiograph features and the respective reports. Furthermore, according to the specificity of data from different domains, a similar mechanism may have different effects in different areas. We are motivated to  explore a different feature fusion pattern apart from the usually adopted cascaded spatial and channel-wise attention mechanism, 

To enhance the visual features obtained above, we employ a stack of m-linear attention blocks, which is capable of catching high-order feature interactions.
Taking the combined regional features $f$ as the initial input keys  ${\bf{K}}=\{{\bf{k}}_i\}_{i=1}^N$ and the values  ${\bf{V}}=\{{\bf{v}}_i\}_{i=1}^N$, and the global features $f_{global}$ as the initial input query $\textbf{Q}$, we attend the input features with our m-linear attention as in Figure~\ref{fig.attention}.

First, we do a low-rank bi-linear pooling  \cite{kim2018bilinear} on both the input query and key, and the query and values to get the joint bi-linear representation of query-key $\textbf{M}^k_i$ and query-value $\textbf{M}^v_i$, which encodes the second-order feature interactions of query-key and query-value.

\vspace{-1.5em}
\begin{small}\label{Eq:Eq1}
	\begin{align}
		{\bf{M}}^k_i&=\sigma\left({W}_k{{\bf{k}}_i}\right) \odot \sigma\left({W}^k_q{\bf{Q}}\right),\\{\bf{M}}^v_i&=\sigma\left({W}_v{{\bf{v}}_i}\right) \odot \sigma\left({W}^v_q{\bf{Q}}\right).
	\end{align}
\end{small}
\vspace{-1.5em}

Next, we transform the query-key representations $\textbf{M}^k_i$ into query-key representations $\textbf{M}^{'k}_i$ with an embedding layer. Both spacial and channel-wise bi-linear attention distributions are obtained according to the transformed query-key representations $\textbf{M}^{'k}_i$. Specifically, we perform  spatial attention via another embedding layer to obtain the spatial attention weights $ a^s$, and normalize it with a softmax layer to get the spatial attention distribution ${\bm{A}}^s$.  Meanwhile, the channel-wise attention is achieved via a squeeze and excitation operation \cite{hu2018squeeze}, in which the squeeze performs a global average pooling to obtain a global channel-wise descriptor $\bar{\bf{M}}$, and the excitation produces the channel-wise attention distribution ${\bm{A}}^c$  with the gating mechanism via a fully connected (FC) layer along with a sigmoid.

\vspace{-1.5em}
\begin{small}\label{Eq:Eq4}
	\begin{align}
			{\bf{M}}^{'k}_i &= \delta\left({W}^k_m{\bf{M}}^k_i\right),\\	a^s_{i}&={W}_s{\bf{M}}^{'k}_i, {\bm{A}}^s=softmax\left({\bf{a}}^s\right),\\
		\bar{\bf{M}}&={1 \over N}\sum\nolimits_{i = 1}^N {{\bf{M}}^{'k}_i},\\
		{\bf{a}}^c&={W}_e \bar{\bf{M}},{\bm{A}}^c=sigmoid\left(\bf{a}^c\right).
	\end{align}
	\vspace{-1.5em}
\end{small}

Finally, we generate the m-linear-attended features $\hat{f}$ by computing the combination of spatial and channel-wise weighted bi-linear feature representation.

\vspace{-1.2em}	
\begin{small}\label{Eq:Eq7}
	\begin{align}
			\hat{f}=&Attention\left({\bf{K}}, {\bf{V}}, {\bf{Q}}\right)=
			\notag
			\\ Concat( &{W}_c ({A}^c\odot\bf{M}^v_i), \sum\nolimits_{i = 1}^N {{A}^s{{\bf{M}}^v_i}}),
		\vspace{-1.5em}
	\end{align}
\end{small}	

\noindent where ${W}_k$,  ${W}_v$, ${W}_q^k $, ${W}_q^v$, ${W}_m^k$, ${W}_s $, ${W}_e $, and ${W}_c $     are embedding matrices, $\sigma (\cdot)$ denotes an \textbf{ELU} unit (Exponential Linear Unit), $\delta (\cdot)$ denotes a  \textbf{ReLU} unit, and $\odot$ represents element-wise multiplication.

In order to extend to higher-order intra-modal feature interactions, our visual encoder is composed of a stack of m-linear attention blocks. 
Specifically, we take the previous output m-linear attended feature  $\hat{f}^{(n-1)}$ as input query Q, along with the current keys ${\bf{K}}^{(n-1)}=\{{\bf{k}}_i^{(n-1)}\}_{i=1}^N$ and values ${\bf{V}}^{(n-1)}=\{{\bf{v}}_i^{(n-1)}\}_{i=1}^N$, which are updated through concatenating with the new attended feature via a residual connection and layer normalization:

\vspace{-1.2em}	
\begin{small}\label{Eq:Eq8-1}
	\begin{align}
	\hat{f}^{(n)}&=Attention\left( {\bf{K}}^{(n-1)}, {\bf{V}}^{(n-1)}, \hat{f}^{(n-1)}\right),\\
	{\bf{k}}_i^{(n)}&=LN(\delta({W}^k_n[\hat{f}^{(n)},{\bf{k}}_i^{(n-1)}])+{\bf{k}}_i^{(n-1)}),\\
	{\bf{v}}_i^{(n)}&=LN(\delta({W}^v_n[\hat{f}^{(n)},{\bf{v}}_i^{(n-1)}])+{\bf{v}}_i^{(n-1)}),
	\end{align}
\end{small}	
\vspace{-1.3em}

\noindent where $W^v_n$ and $W^k_n$ are embedding matrices, $\bf{v}_i^{(n)}$ and $\bf{k}_i^{(n)}$ represents the $i^{th}$ key/value, $\delta (\cdot)$ denotes a \textbf{ReLU} unit, 
and $LN$ denotes layer normalization. We repeat this process four times to achieve $8^{th}$-order feature interactions.

\vspace{-1em}
\subsection{Sentence Decoder RNN}

The sentence decoder takes the combination of different order attended region-level visual features $\hat{f}^{(n+1)}$ and the global image feature $f_{global}$ as input $\tilde{f}$ , and generates the report word by word. In order to exploit high-order inter-modal interactions between visual features and semantic features, we employ an m-linear attention-based LSTM decoder: 

\vspace{-1.2em}	
\begin{small}\label{Eq:Eq9}
	\begin{align}	
	\tilde{f}&={\rm{W}}_c[\hat{f}^{(0)},\hat{f}^{(1)},...,\hat{f}^{(1+N)}],\\
	h_t &= LSTM(\tilde{f}; w_{t}, h_{t-1}, c_{t-1}),
	\end{align}
\end{small}	
\vspace{-1.3em}


\noindent where ${\rm{W}}_c$ is an embedding matrix,  $\hat{f}^{(0)} = f_{global}$,  $w_{t}$ is the current input word, $h_{t-1}$ is the previous LSTM hidden state, and $c_{t-1}$ is the previous context vector. Next, we take the output hidden state $h_t$ of the LSTM as the input query \textbf{Q} of an m-linear attention block, and the final output attended feature $f^{(N+1)}$ from the visual encoder as keys \textbf{K} and values~\textbf{V}:

\vspace{-1.2em}	
\begin{small}\label{Eq:Eq11}
	\begin{align}
	\hat{f}_d=&Attention\left({\bf{K}}, {\bf{V}}, {\bf{Q}}\right)\\=&
		\notag
		Attention\left({f^{(N+1)}}, {f^{(N+1)}}, {h_t}\right).
		\vspace{-1.5em}
	\end{align}
\end{small}

After that, we compute the current context vector $c_t$ with a residual connection and a gated linear unit (GLU), followed by a fully connected layer: 

\vspace{-1.2em}	
\begin{small}
	\begin{align}
		c_t = {\rm{W}}_c(\epsilon (h_t + {\rm{W}}_L(h_t+\hat{f}_d))),
	\end{align}
\end{small}	
\vspace{-1.3em}

\noindent where ${\rm{W}}_c$ and  ${\rm{W}}_L$   are embedding matrices, and $\epsilon (\cdot)$ denotes a \textbf{GLU} unit. Finally, the context vector $c_t$ is sent into a softmax layer to predict the next word $w_{t+1}$:

\vspace{-1.2em}	
\begin{small}\label{Eq:Eq12}
	\begin{align}
		w_{t+1} = Softmax(c_t).
	\end{align}
	\vspace{-1.5em}
\end{small}

\vspace{-1em}
\subsection{Repetition Penalty Module}

Different from traditional image captioning tasks, which aim at generating a single sentence, our task requires the generation of paragraphed reports, which consist of hundreds of words. Inevitably, this leads to the increase of generation difficulty, thus resulting in  repeated terms. Furthermore, the employed metrics (i.e., BLEU-1), which focus on the matching of words, also aggravate this problem. Though our proposed m-linear attention alleviates the problem to some extent by improving the capacity of feature extraction, we also propose a repetition penalty module that constrains the probabilities of words resulting in repeated trigrams to decrease the probability of repeated terms generated in the reports.

 \cite{melas2018training} also discovered this problem and proposed to add a penalty when generating words. However, it only assigns a selected weight to the penalty while not giving the process of finding the weight. Furthermore, according to the difference of datasets, it is hard to assure the robustness of the penalty. Since the more the occurrence of repetition terms, the less likely the sentence to be correct and readable, 
 in order to address the above-mentioned problems, we propose to assign an exponential weight to the generated repeated words.
Specifically, we update the log-probability of the output word by subtracting a value proportional to the number of times the trigram has been generated:

%
\vspace{-1.2em}
\begin{small}
	\begin{align}
		p_w = p_w - (1 - e^{-n_w}), 
	\end{align}
\end{small}
\vspace{-1.3em}

\noindent where $p_w$ is the log-probability of the word $w$, and $n_w$ is the number of times that the trigram has generated the word $w$.
We employ this update mechanism during our greedy search process in SCST in order to generate more diverse paragraphs, thus avoiding the repetition problem.

\vspace{-1em}
\subsection{Reinforcement Learning for Comprehensive Optimization}
The reinforcement learning algorithm commonly used in existing medical report generation 
\cite{10.1007/978-3-030-32692-0_77,liu2019clinically} is the self-critical sequence training (SCST) algorithm \cite{rennie2017self}, which directly optimizes the automatic natural language generation (NLG) metrics. 
Specifically, SCST adopts a policy gradient method to optimize a non-differentiable metric such as CIDEr. In order to normalize the reward and reduce the variance during training, SCST utilizes the REINFORCE algorithm with a baseline, which is obtained from the inference reward by greedy search. The goal is to minimize the negative expected reward.
The final gradient of the optimization object is:

%

\vspace{-1em}
\begin{small}
	\begin{align}
		\frac{\partial{L(\theta)}}{\partial{s_t}} = (r(w^s) - r(\hat{w}))\bigtriangledown_{\theta}\log p_{\theta}(w^s | x), 
	\end{align}
	\vspace{-1em}
\end{small}

\noindent where $w^s  = (w_1^s, . . . ,w_T^s )$ is a Monte-Carlo sample from~$p_{\theta}$, $p_{\theta}$ is our generation model,  $r(w^s)$ is the current reward, and $r(\hat{w})$ is the reward obtained by the inference algorithm. 

\renewcommand{\algorithmicrequire}{\textbf{Input:}}
\renewcommand{\algorithmicensure}{\textbf{Output:}} 

\setlength{\textfloatsep}{0pt}
\begin{algorithm}[!t]
	\caption{Search Solution 1} 
	\label{search1} 
	\begin{algorithmic}
		\REQUIRE $A$: a set of weights of all the metrics \\ 
	    	$n$: the search space\\
	    	$ Score $: model of compute the NLP metric scores \\
	    	$S$: sum of scores of each weight combination 
		\ENSURE $A$: a set of searched weights of all the metrics
		\textbf{Initialize} $A_i=1$ ; $n=5$ ; 
		\WHILE{existing the most influenced $A_i$ hasn't been searched}
		\FOR {each $i \in [1,7]$ and $A_i$ hasn't been searched }
		\STATE $A_i  \gets  A_i+1$
		\STATE $S_i \gets Score(A)$
		\ENDFOR
		\STATE find out the most influenced $A_i$ based on $S$ 
		\STATE set other element of $A$ to be 1 
		\FOR { each $j \in [2,5]$}
		\STATE $S_{i,j} \gets Score(A) $
		\STATE $ A_{i,j} \gets A_{i,j-1} + 1$
		\ENDFOR
		\STATE find out the most influenced $A_{i,j}$ based on $S$ 
		\STATE $A_{i} \gets A_{i,j} $ 
	\ENDWHILE
	\end{algorithmic} 
\end{algorithm}

Since most of the existing reinforcement learning-based methods take CIDEr  \cite{vedantam2015cider} as the reward, it inevitably leads to the failure of optimization on other metrics. Furthermore, the existing works did not reasonably explain the use of the CIDEr-based reward, since the success of the CIDEr-based reward was proved in the task of image-captioning, while it has not been verified in the area of medical report generation.
Intuitively, we consider verifying the effectiveness of all the metrics and utilizing them to achieve an overall promotion. 
However, since each metric focuses on different aspects, and some metrics are more important in our task (i.e., METEOR and ROUGE-L further consider the recall besides precision compared with BLEU), simply applying the same weight to each metric is not reasonable. We are motivated to attach different weights to each metric to achieve a weighted optimization.
\cite{8237362} also proposed to adopt a mixture of the metrics, while they did not give any explanation for the combination or provide a solution to find the suitable weight. 
We propose a different weighted hybrid reward and offer a solution to search for the best weight. 


In our work, the seven most frequently used natural language generation metrics are adopted as the rewards. We propose to use a different weighted hybrid reward to achieve an overall promotion. However, the search for the optimal weight is time-consuming. The simplest way is to apply grid search, whose complexity is $O(n^7)$. We propose to employ a greedy-search-based solution.  
   Assigning the weight of one to each metric, we are supposed to find the most influential metric and increase by one each time in the search space of $n$ until we find the best weight for it. After that, fixing the weight for that metric, we repeat the above process for the other metrics, until we find the best combination. In this way, we figure out that the complexity of the solution is $O(7+(n-1) + 6 + (n-1) +...+2 +(n-1) + (n-1)) = O(20+7n)$, which reduces the complexity from an exponential level to a linear level.

    
The gradient of our optimization object is: 
 \vspace{-0.5em}
\begin{small}
	\begin{align}
		\frac{\partial{L(\theta)}}{\partial{s_t}} = (\sum_{i=0}^{7} \lambda_i r(w^s) - \sum_{i=0}^{7} \lambda_i r(\hat{w}))\bigtriangledown_{\theta}\log p_{\theta}(w^s | x), 
	\end{align}
	\vspace{-1em}
\end{small}

\noindent where $\lambda_i$ is the weight of the corresponding metric.

\vspace{-0.5em}
\section{Experiments}
\label{sec4}

\begin{table*}[!t] 
\vspace{-1.5em}
	\caption{Experimental results of HReMRG-MR and the state-of-the-art baselines on IU X-Ray (upper part) and MIMIC-CXR (lower part).
	}
	\centering   
	
	\begin{tabular}{l|l|cccccccc}
	
		\hline  
		Dataset & Model & BLEU-1 & BLEU-2 & BLEU-3 & BLEU-4 &CIDEr  & METEOR & ROUGE-L   & Score  \\ \hline 
		\multirow{5}{*}{\scriptsize{\textbf{IU X-Ray}}} 
		& top-down \cite{anderson2018bottom} & 0.279 & 0.178 & 0.119 & 0.079 & 0.206 & 0.144 & 0.334 &1.339   \\ 
		&MRMA \cite{xue2018multimodal} & 0.382 & 0.252 & 0.173 & 0.120 & 0.325 &  0.163 & 0.309 &1.724  \\ 
		&RTMIC  \cite{10.1007/978-3-030-32692-0_77} & 0.3356& 0.2236&0.15&0.1003&0.2779&0.1664&0.3371&1.5909 \\
		&X-LAN \cite{Pan2020} & 0.3782 & 0.2636 & 0.1858 & 0.1314 & 0.4508 & 
		0.1735 & 0.3490 & 1.8004 \\
		& HReMRG-MR  & \textbf{0.4399} & \textbf{0.3059} & \textbf{0.2139} & \textbf{0.1490} & \textbf{0.5239} & \textbf{0.1971} & \textbf{0.3810} & \textbf{2.2107} \\
		
		\midrule 
		\multirow{5}{*}{\scriptsize{\textbf{MIMIC-CXR}}}  
		& top-down \cite{anderson2018bottom} & 0.233 & 0.159 & 0.119 & 0.093 &0.359&  0.134 &  0.319 &1.416\\ 
		&MRMA \cite{xue2018multimodal}  &0.361 & 0.244 & 0.182 & 0.141& 0.324 & 0.157 & 0.330 &1.739  \\
		&RTMIC \cite{10.1007/978-3-030-32692-0_77} &0.3654&0.2448&0.1772&0.134&0.3575&0.1521&0.3208&1.7518 \\
		&X-LAN \cite{Pan2020} &0.362&0.2579&0.1801&0.126&0.365&0.169&0.3402&1.8002\\
		& HReMRG-MR & \textbf{0.4806} & \textbf{0.3431}&\textbf{0.2555} & \textbf{0.1921} & \textbf{0.3715} & \textbf{0.2070} &  \textbf{ 0.3802} & \textbf{2.2301} \\
		
		\hline 
	\end{tabular} 
	\vspace*{-1em}
	\label{Table:baseline} 
\end{table*}

\begin{figure*}[!t] 
	\centering {\includegraphics[width=0.92\textwidth]{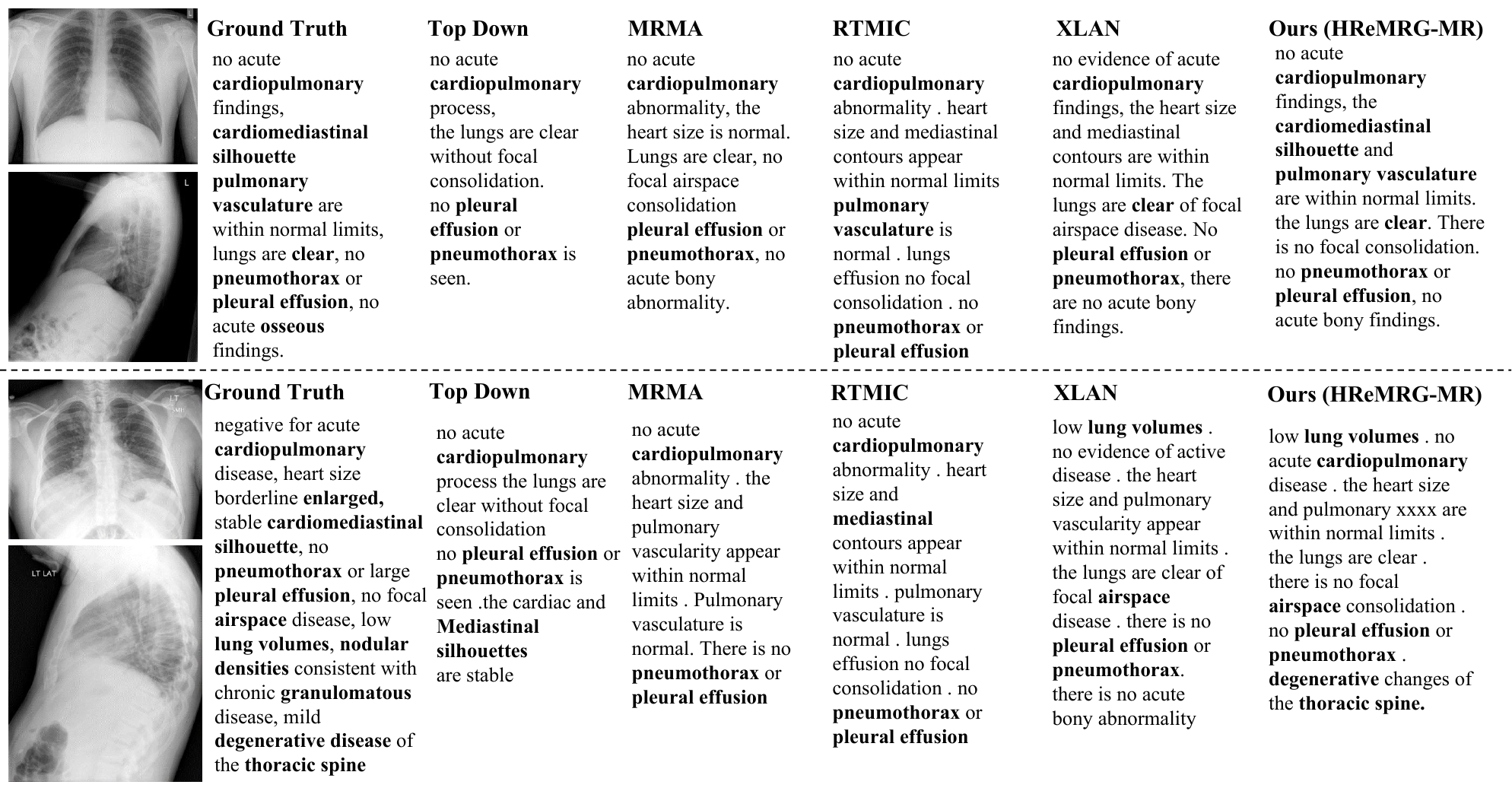}}
	\vspace{-0.5em}
	\caption{Examples of reports generated by Top Down, MRMA, RTMIC, XLAN, and HReMRG-MR.}
	\label{fig.sample}
	\vspace*{-1.5em}
\end{figure*}

\subsection{Datasets}

In order to evaluate the performance of our proposed ReMRG-MR, extensive experiments have been conducted on the following two public chest x-ray image datasets.

\smallskip \noindent \textbf{IU X-Ray} \cite{demner2016preparing} is a set of chest x-ray images with paired diagnostic reports collected by Indiana University. It contains 7,470 medical images and 3,955 corresponding reports, which mainly consist of two sections: \textit{Impression} and \textit{Findings}. As some images or reports are missing, 
we filtered out 6,222 images including both 
frontal and lateral views, and 3,111 paired reports.
We take  \textit{Impression} and \textit{Findings} as our generation object and preprocess the texts by converting all words to lowercase. After that, we tokenize the reports, resulting in 2068 unique words in total. As most words occur rarely, we exclude tokens appearing less than 5 times and get 776 tokens.

\smallskip \noindent\textbf{MIMIC-CXR} \cite{johnson2019mimic} is the largest publicly available dataset of chest medical images with free-text radiology reports, which consists of 377,110 images along with 227,827 free-text  reports. Similarly, we filtered out 76,724 pairs of images combined of posteroanterior (PA) or anteroposterior (AP) view and lateral (LL) view with their reports, which consist of both \textit{Impression} and \textit{Findings}.
We then apply the same preprocessing and keep tokens with more than 5 occurrences, ending up with 2991 tokens overall. 

Finally, for both datasets, we randomly partition them by patients into training, validation, and testing with a  ratio of 7:1:2 and make sure that there is no overlap between the sets. Words excluded are replaced by the token of `UNK'.

\vspace{-1em}
\subsection{Evaluation Metrics}

To evaluate the performance of our proposed model, we use several different automatic language generation metrics including BLEU \cite{papineni2002bleu}, METEOR \cite{banerjee2005meteor}, ROUGE-L \cite{lin2004rouge}, and CIDEr \cite{vedantam2015cider}. In particular, BLEU is a popular machine translation evaluation metric, which aims at measuring the ratio of correct matches of n-grams between generated sequences and the ground truth.
To address some defects in BLEU, METEOR further considered the accuracy and recall based on the whole corpus.
ROUGE-L is a metric for summary evaluation via measuring the longest common sequence that matches between two sequences based on recall. 
Differently from the aforementioned metrics, CIDEr was designed especially for image description, which is most suitable for our task. It computes the term frequency-inverse document frequency (TF-IDF) weight of n-grams to obtain the similarity between the candidate sequences and the reference sequences.

\vspace{-1em}
\subsection{Baselines}

On both datasets, we compare our method with four state-of-the-art image captioning and medical report generation models: 
(i) our re-implementation of the top-down model \cite{anderson2018bottom}, which is a classic encoder-decoder-based model for image captioning employing a conventional attention mechanism that calculates the contribution of regional features to the texts to be generated, and  
(ii) MRMA \cite{xue2018multimodal}, an encoder-decoder-based model specially designed for medical report generation, in which reports are generated sentence by sentence with a recurrent way to generate long paragraphs. 
(iii) RTMIC, which is a state-of-the-art medical report generation method based on reinforcement learning, enhancing the capacity of the generation model with reinforcement learning, and increasing the clinical accuracy with a transformer. 
(iv) X-LAN \cite{Pan2020}, which is an image captioning model employing x-linear attention and improving it with reinforcement learning taking CIDEr as the reward. As image captioning is similar to our task to some extent, we also take this model as our baseline. 
For our implemented methods, we use the same visual features and train/val/test split on both datasets.



\begin{table*}[!t] 
	\caption{
	Ablation studies, where X indicates x-lienar attention, M indicates m-linear attention, and R indicates repetition penalty. 
	}
	\centering   
	\begin{tabular}{l|l|cccccccc}
		\hline  
		Dataset & Model & BLEU-1 & BLEU-2 & BLEU-3 & BLEU-4 &CIDEr  & METEOR & ROUGE-L   & Score  \\ \hline 
		\multirow{5}{*}{\scriptsize{\textbf{IU X-Ray}}} 
		&HReMRG & 0.3491 & 0.2245 & 0.1551& 0.1139&0.4489&0.1511&0.2925 & 1.7351\\
		&HReMRG-X&\textbf{0.4399}& 0.3081& 0.2135 & 0.1466 & 0.4378 & 0.1942 & 0.374 & 2.1141\\
		& HReMRG-XR & 0.4374 & \textbf{0.3096 }& \textbf{0.2177}& \textbf{0.1544 }& 0.4589 & 0.1926 & 0.3733 & 2.1439 \\
		& HReMRG-M & 0.4322 & 0.3034 & 0.2113 & 0.1462 & 0.4929 &0.1945 &  0.3795 & 2.1599 \\
		& HReMRG-MR & \textbf{0.4399} & 0.3059 & 0.2139 & 0.1490 & \textbf{0.5239} & \textbf{0.1971} & \textbf{0.3810} & \textbf{2.2107} \\
		
		\midrule 
		\multirow{5}{*}{\scriptsize{\textbf{MIMIC-CXR}}}  
		& HReMRG& 0.3084 & 0.2131 & 0.1611 & 0.1252 & 0.3164 & 0.1461 & 0.3383 & 1.6086 \\
		& HReMRG-X & \textbf{0.4849}& 0.3429 & 0.2548 & 0.1897 & 0.3423 &\textbf{ 0.2092} &0.3742 & 2.1980\\
		&HReMRG-XR& 0.4760 &  0.3385  &0.2533 & 0.1908 & 0.3541&  0.2059 & \textbf{0.3910}&  2.2096\\
		& HReMRG-M & 0.4821&	0.3428&	\textbf{0.2558} &	0.1920&	0.3529 &0.2081 &	0.3752 &	2.2089 \\
		& HReMRG-MR & 0.4806 & \textbf{0.3431}&0.2555 & \textbf{0.1921} & \textbf{0.3715} & 0.2070 &   0.3802 & \textbf{2.2301} \\
		
		\hline 
	\end{tabular} 
	
	\vspace*{-1em}
	\label{Table:abalationStudy} 
\end{table*}

\begin{figure*}[!t] 
	\centering {\includegraphics[width=0.92\textwidth]{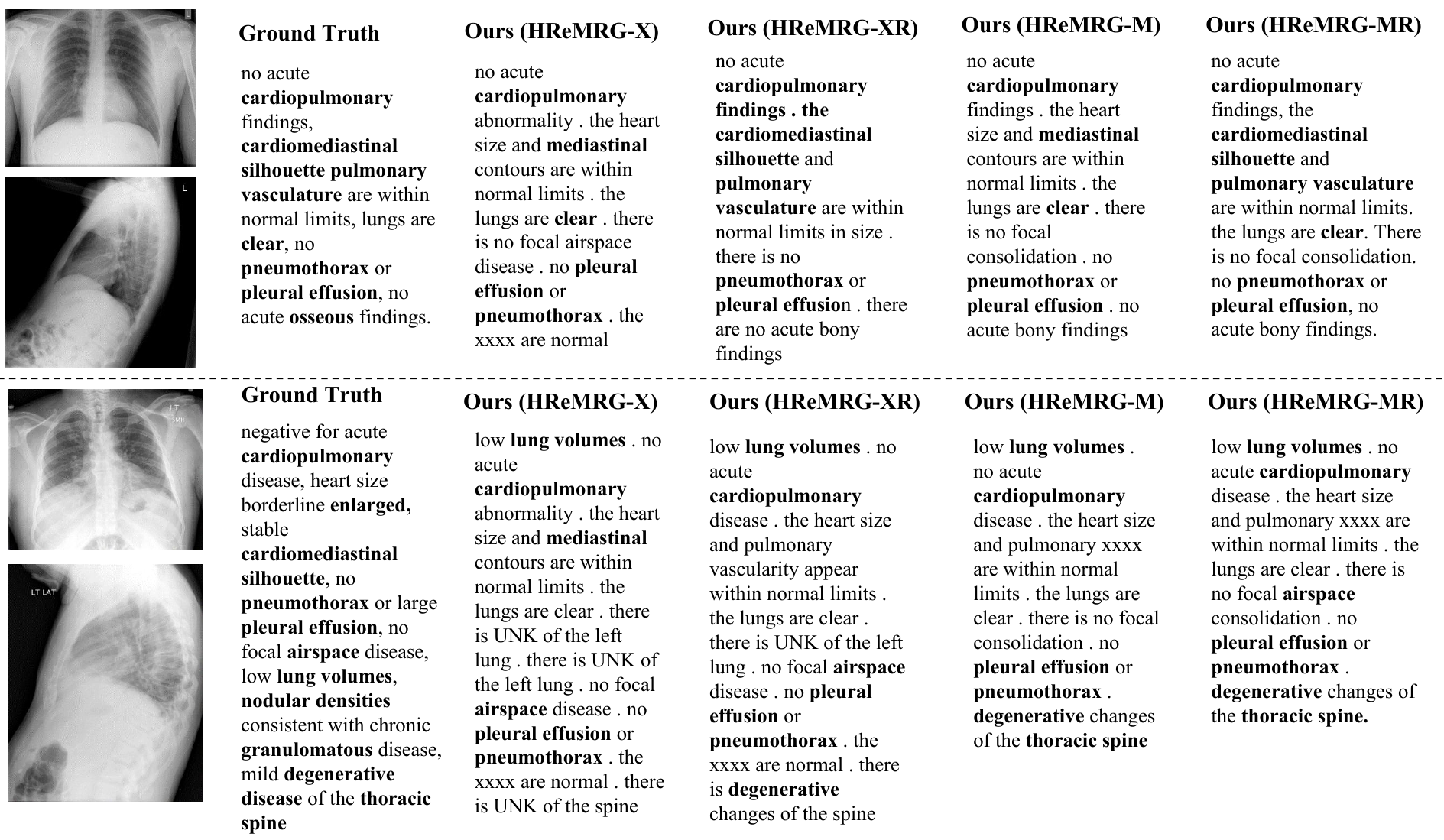}}
	\caption{Examples of reports generated by HReMRG, HReMRG-X, HReMRG-XR, HReMRG-M, and HReMRG-MR.}
	\label{fig.sample-2}\vspace*{-1em}
\end{figure*}

\vspace{-1em}
\subsection{Implementation Details}
We use  ResNet-101 pre-trained on ImageNet \cite{deng2009imagenet} to extract the region-level features, which are from the last convolutional layer. As both views of medical images are sent into the model simultaneously, two original 2048-dimensional region features are concatenated to a 4096-dimensional vector, which is later transformed into the visual embedding of size 1024.
Next, four stacks of m-linear attention blocks are used to explore the high-order intra-modal interactions.
During \textit{Decode}, we set the hidden layer size and word embedding dimension to 1024.
As most reports are within a fixed length, we set the max generation length of IU X-Ray and MIMIC-CXR to 114 and 184, respectively.

In training, we first pre-train the model with cross-entropy loss for 60 epochs with a batch size of 8 with NVIDIA RTX 2080 GPUs. Setting the base learning rate to  0.0001, we utilize the Noam decay strategy with 10,000 warmup steps with Adam \cite{kingma2014adam} as the optimizer.
In the reinforcement learning stage, we set the different weighted hybrid reward obtained via our search solution  as our training reward. The search space k is set to be 5. We set the base learning rate to 0.00001 and decay with CosineAnnealing with a period of 15 epochs. We also set the maximum iteration to  60 epochs and the batch size to 3. We use beam search with a beam size of 2 for training.

\vspace{-1em}
\subsection{Main Results}

Table~\ref{Table:baseline} shows the experimental results of the proposed HReMRG-MR and four baselines in terms of seven natural language generation metrics and the sum of them, where all baselines are re-implemented by ourselves. In addition, Figure~\ref{fig.sample} exhibits some examples of reports generated by these models. 

Generally, our proposed HReMRG-MR, which employs different weighted hybrid reward and m-linear attention with repetition penalty utilized, greatly outperforms all state-of-the-art baselines in all metrics in Table~\ref{Table:baseline}. From Figure~\ref{fig.sample}, it can be observed that HReMRG-MR also generates more coherent and accurate reports than the baselines.

Particularly, MRMA outperforms top-down in almost all metrics, since MRMA adopts a recurrent text generation method in addition to an attention mechanism, which is specially designed for paragraphed reports generation. Compared to the classic encoder-decoder-based model top-down, the performance of RTMIC has an obvious advantage for the utilization of reinforcement learning. Comparing RTMIC and MRMA, they achieve similar scores in terms of overall performance. It proves that the improvement of long text generation and the use of reinforcement learning are both effective improvement methods. When it comes to the performance of X-LAN, it proves the effectiveness of x-linear attention, since they both adopt reinforcement learning. Furthermore, it also outperforms MRMA, which proves that the utilization of a suitable attention mechanism and the employment of reinforcement learning is more effective than the recurrent text generation method. Therefore, our research work is conducted on this basis. 


\begin{table*}[!t]  
	\caption{
	Results of using different rewards in Reinforcement learning,	
	where SWHR denotes same weighted hybrid reward, and DWHR denotes different weighted hybrid reward. Best results are highlighted and second best results are underlined.
	}
	\centering   
	\begin{tabular}{l|l|cccccccc}
		\hline  
		Dataset & Reward & BLEU-1 & BLEU-2 & BLEU-3 & BLEU-4 &CIDEr  & METEOR & ROUGE-L   & Score  \\ \hline 
		\multirow{9}{*}{\scriptsize{\textbf{IU X-Ray}}} 
		& BLEU-1 & \underline{0.4420} & 0.2968 & 0.2038 & 0.1397 & 0.3423 & 0.1887 & 0.3597 & 1.9732 \\ 
		& BLEU-2 & 0.4225 & \underline{0.3039} & \underline{0.2128} & 0.1491 & 0.4292 & 0.1874 & 0.3633 & 2.0683 \\
		& BLEU-3 & 0.4034 & 0.2926 & 0.2100 &\textbf{0.1510} & 0.4348 &
		0.1832 &  0.3630 & 2.0436 \\
		& BLEU-4 & 0.3762 & 0.2772 & 0.2041 & 0.1488 & 0.3944 & 0.1670 & 0.3241 & 1.8917 \\
		& CIDEr & 0.3782 & 0.2636 & 0.1858 & 0.1314 & \textbf{0.4508} & 
		0.1735 & 0.3490 & 1.8004 \\
		& METEOR & 0.4377 & 0.2850 & 0.2000 & 0.1400 & 0.2021 & \underline{0.1915} & 0.3442 & 1.9499 \\
		& ROUGE-L & 0.3632 & 0.2717 & 0.1931 & 0.1357 & 0.4147 & 0.1729 & \underline{0.3716 }& 1.9311 \\
		
		\cline{2-10}   
		& SWHR & 0.4160 & 0.2952 & 0.2101 & \underline{0.1497} & \underline{0.4507} & 0.1870 & 0.3650 & \underline{2.0741} \\
		& DWHR &\textbf{0.4399}& \textbf{0.3081}& \textbf{0.2135} & 0.1466 & 0.4378 & \textbf{0.1942} &\textbf{ 0.374} & \textbf{2.1141}\\
		
		\midrule 
		\multirow{9}{*}{\scriptsize{\textbf{MIMIC-CXR}}}  
		& BLEU-1 & \textbf{0.5002} & 0.3205 & 0.2239 & 0.1630 & 0.2674 & 0.2072 & 0.3632 & 2.0454  \\ 
		& BLEU-2 & 0.4801 & \underline{0.3415} & 0.2523 & 0.1870 & 0.3476 & 0.2066 & 0.1698 & 2.1850 \\
		& BLEU-3 & 0.4713 & 0.3388 & \underline{0.258}& \underline{0.1942} & 0.3456 & 0.2036 & 0.1303 & 2.1800 \\
		& BLEU-4 & 0.4613&  0.3373&\textbf{ 0.2601}&  \textbf{0.2033}&  0.3494& 0.1998&  \underline{0.381}&  2.1922\\
		&CIDEr & 0.362&0.2579&0.1801&0.126&\underline{0.365}&0.169&0.3402&1.8002\\
		& METEOR & 0.3972 & 0.2835 & 0.2148 & 0.1658 & 0.1166 & \textbf{0.2183} & 0.3596 & 1.7558\\ 
		& ROUGE-L & 0.4150 & 0.2924 & 0.2185 & 0.1648 & 0.3323 & 0.1822 & \textbf{0.3927} & 1.9979\\
		
		\cline{2-10}
		& SWHR & 0.4699 & 0.3339 & 0.2504 & 0.1910 & \textbf{0.3753}& 0.2043 & 0.3707 & \underline{2.1956} \\
		& DWHR & \underline{0.4849 }& \textbf{0.3429} & 0.2548 & 0.1897 & 0.3423 & \underline{0.2092} &0.3742 & \textbf{2.1980}\\
		\hline 
	\end{tabular} 
	
	\vspace*{-1em}
	\label{Table:report-reward} 
\end{table*}

\vspace{-1em}
\subsection{Ablation studies}

In this section, we report on a series of ablation experiments to demonstrate the effectiveness of the proposed model. Specifically, we additionally implemented four versions of our proposed HReMRG-MR by: 
(i) using only different weighted hybrid reward for medical report generation, called HReMRG, 
(ii) using different weighted hybrid reward and x-linear attention for medical report generation, called HReMRG-X,
 (iii) then adding repetition penalty to get another version of the model, called HReMRG-XR, and 
(iv) using different weighted hybrid reward and m-linear attention for medical report generation, called HReMRG-M.


In Table~\ref{Table:abalationStudy}, we can see that with x-linear attention utilized, the performance of HReMRG-X boosts a lot compared with HReMRG, and when replacing x-linear attention with m-linear attention, the scores of HReMRG-M further improve. On the basis of HReMRG-M, we utilize the repetition penalty to obtain HReMRG-MR, which achieves the highest scores among all the versions of models.

In Table~\ref{Table:abalationStudy}, we use HReMRG-X and HReMRG-XR as the baselines to further analyze the effectiveness of m-linear attention, because they are equivalent to HReMRG-M and HReMRG-MR if their attention mechanisms are replaced by m-linear attention.
As shown in Table~\ref{Table:abalationStudy} HReMRG-M (resp.,  HReMRG-MR) outperforms HReMRG-X (resp., HReMRG-XR) on most metrics especially on CIDEr, which incorporates TF-IDF of words in the entire evaluation corpus, thus giving higher weight to informative phrases. This is because the method fusing channel-wise attention and spatial attention suits our task and datasets better, so that m-linear attention can better explore high-order feature interactions and achieve multi-modal reasoning, making it possible to generate more accurate reports.
Examples in Figure~\ref{fig.sample-2} tell the same story: the description generated by  HReMRG-MR is more accurate (i.e., has more correctly matched terms) than that generated by HReMRG-XR.


HReMRG-X and HReMRG-M are used as the baselines to further evaluate the effectiveness of repetition penalty, because they are equivalent to HReMRG-XR and HReMRG-MR if the repetition penalty is introduced to them.
As shown in Table~\ref{Table:abalationStudy} HReMRG-XR (resp.,  HReMRG-MR) achieves much better performances than HReMRG-X (resp.,  HreMRG-M) in terms of most metrics,  especially on BLEU-4 and CIDEr. This is because, with the repetition penalty utilized, the texts generated tend to generate more diverse phrases avoiding repetitive words with little information. Thus, the scores of BLEU-4, which evaluate the matching of four-gram phrases, raise obviously, and the scores of CIDEr, which measures the matching of informative phrases, raise accordingly. Similarly, the example report of HReMRG-MR in Figure~\ref{fig.sample} is also more accurate and readable than that generated by HReMRG-M.
These findings both demonstrate that by applying the repetition penalty, medical report generation models can achieve much better performances.
Hence, we employ m-linear attention and repetition penalty in our final model HReMRG-MR.

%
%
%

\begin{table*}[!t] 
	\small
	\vspace{-1em}
	\caption{Proposed Search solution for three weighted hybrid rewards on IU X-Ray
	}
	\centering \vspace*{-1ex}  
	\begin{tabular}{l|l|cccccccc}
		\hline  
		Dataset & Reward & BLEU-1 & BLEU-2 & BLEU-3 & BLEU-4 &CIDEr  & METEOR & ROUGE-L   & Score  \\ \hline 
		\multirow{8}{*}{\scriptsize{\textbf{IU X-Ray}}} 
		& 0:0:0:1:1:1:0 & 0.4210 & 0.2968 & 0.2074 & 0.1457 & 0.4114 & 0.1868 & 0.3624 & 2.0314 \\
		\cline{2-10}
		& 0:0:0:2:1:1:0 & 0.4054 & 0.2936 & 0.2135 & 0.1551 & 0.3737 & 0.1832 & 0.3534 & 1.9779 \\
		& 0:0:0:1:2:1:0 &  0.3972 & 0.2871 & 0.2068 & 0.1483 & 0.3968 & 0.1835 & 0.3599 & 1.9795 \\
        & 0:0:0:1:1:2:0 & 0.4202 & 0.2958 & 0.2108 & 0.1517 & 0.448 & 0.1888 & 0.3679 & \textbf{2.0833} \\
        \cline{2-10}
        & 0:0:0:2:1:2:0 & 0.4202 & 0.2958 & 0.2108 & 0.1517 & 0.448 & 0.1879 & 0.3647 & 2.0792 \\
        & 0:0:0:1:2:2:0 & 0.4028 & 0.2917 & 0.2106 & 0.1513 & 0.4339 & 0.1828 & 0.3564 & 2.0296 \\ 
        \cline{2-10}
        & 0:0:0:3:1:2:0 & 0.4235 & 0.2997 & 0.2124 & 0.1498 & 0.3959 & 0.1892 & 0.3638 & 2.0343 \\
        \cline{2-10}
        & 0:0:0:1:3:2:0 & 0.4215 & 0.2921 & 0.2081 & 0.1488 & 0.4335 & 0.1853 & 0.3651 & 2.0545 \\
      
		\hline 
	\end{tabular} 
	
	\vspace*{-1em}
	\label{Table:compare-search-proposed} 
\end{table*}

\begin{table*}[!t] 
\vspace{-1em}
\vspace*{2ex}	
	\small
	\caption{Generally Search solution for three weighted hybrid rewards on IU X-Ray
	}
	\centering   \vspace*{-1ex}
	\begin{tabular}{l|l|cccccccc}
		\hline  
		Dataset & Reward & BLEU-1 & BLEU-2 & BLEU-3 & BLEU-4 &CIDEr  & METEOR & ROUGE-L   & Score  \\ \hline 
		\multirow{27}{*}{\scriptsize{\textbf{IU X-Ray}}} 
		& 0:0:0:1:1:1:0 & 0.4210 & 0.2968 & 0.2074 & 0.1457 & 0.4114 & 0.1868 & 0.3624 & 2.0314 \\
		\cline{2-10}
		& 0:0:0:2:1:1:0 & 0.4054 & 0.2936 & 0.2135 & 0.1551 & 0.3737 & 0.1832 & 0.3534 & 1.9779 \\
		& 0:0:0:1:2:1:0 &  0.3972 & 0.2871 & 0.2068 & 0.1483 & 0.3968 & 0.1835 & 0.3599 & 1.9795 \\
        & 0:0:0:1:1:2:0 & 0.4202 & 0.2958 & 0.2108 & 0.1517 & 0.448 & 0.1888 & 0.3679 & \textbf{2.0833} \\
   
        & 0:0:0:2:1:2:0 & 0.4202 & 0.2958 & 0.2108 & 0.1517 & 0.448 & 0.1879 & 0.3647 & 2.0792 \\
        & 0:0:0:1:2:2:0 & 0.4028 & 0.2917 & 0.2106 & 0.1513 & 0.4339 & 0.1828 & 0.3564 & 2.0296 \\ 

        & 0:0:0:3:1:2:0 & 0.4235 & 0.2997 & 0.2124 & 0.1498 & 0.3959 & 0.1892 & 0.3638 & 2.0343 \\
 
        & 0:0:0:1:3:2:0 & 0.4215 & 0.2921 & 0.2081 & 0.1488 & 0.4335 & 0.1853 & 0.3651 & 2.0545 \\

        & 0:0:0:1:3:1:0 & 0.3828 & 0.2781 & 0.2002 & 0.1433 & 0.4573 & 0.1771 & 0.3597 & 1.9984 \\
        
  & 0:0:0:3:1:1:0 & 0.4058 & 0.2956 & 0.2139 & 0.1538 & 0.3695 & 0.1785 & 0.3212 & 1.9383 \\

& 0:0:0:2:2:1:0 & 0.3952 & 0.292 & 0.2116 & 0.1511 & 0.3993 & 0.1792 & 0.3613 & 1.9897 \\
& 0:0:0:2:3:1:0 & 0.3988 & 0.2924 & 0.211 & 0.1508 & 0.3694 & 0.1784 & 0.3499 & 1.9507 \\
& 0:0:0:1:1:3:0 & 0.4048 & 0.2871 & 0.2025 & 0.1428 & 0.447 & 0.1841 & 0.3594 & 2.0277 \\
& 0:0:0:3:2:1:0 & 0.4002 & 0.2916 & 0.2102 & 0.1499 & 0.4215 & 0.1818 & 0.3664 & 2.0216 \\ 
& 0:0:0:3:3:1:0 & 0.3858 & 0.2773 & 0.2009 & 0.1442 & 0.4196 & 0.1753 & 0.3573 & 1.9604 \\
& 0:0:0:2:1:3:0 & 0.4269 & 0.3019 & 0.2131 & 0.1507 & 0.4287 & 0.1894 & 0.3616 & 2.0723 \\
& 0:0:0:1:2:3:0 & 0.4233 & 0.2950 & 0.2062 & 0.1443 & 0.4428 & 0.1895 & 0.3652 & 2.0662 \\
& 0:0:0:3:1:3:0 & 0.4014 & 0.286 & 0.2059 & 0.1483 & 0.474 & 0.1828 & 0.3584 & 2.0567\\
& 0:0:0:2:2:3:0 & 0.3134 & 0.2971 & 0.2134 & 0.1534 & 0.3633 & 0.1836 & 0.3333 & 1.9575 \\
& 0:0:0:2:3:3:0 & 0.4048 &  0.2842 & 0.2019 & 0.1446 & 0.4427 & 0.1843 & 0.3591 & 2.0216 \\
& 0:0:0:1:3:3:0 & 0.3983 & 0.2854 & 0.2022 & 0.1443 & 0.452 & 0.1805 & 0.3573 & 2.02 \\
& 0:0:0:3:2:3:0 & 0.4215 & 0.296 & 0.2067 & 0.145 & 0.4289 & 0.1881 & 0.3607 &  2.047 \\
& 0:0:0:3:3:3:0 & 0.3998 & 0.2827 & 0.2077 & 0.1436 & 0.4195 & 0.1824 & 0.3569 & 1.9857 \\
& 0:0:0:2:2:2:0 & 0.4043 & 0.2998 & 0.217 & 0.157 & 0.4298 & 0.1829 & 0.3564 & 2.0472 \\
& 0:0:0:2:3:2:0 & 0.4037 & 0.2902 & 0.2069 & 0.1496 & 0.4246 & 0.1818 & 0.3564 & 2.0132 \\
& 0:0:0:3:2:2:0 & 0.3958 & 0.2844 & 0.2065 & 0.1490 & 0.4749 & 0.1766 & 0.3589 & 2.046 \\
& 0:0:0:3:3:2:0 & 0.41 & 0.2963 & 0.21 & 0.1479 & 0.3906 & 0.1853 & 0.3587 & 1.9988 \\

		\hline 
	\end{tabular} 
	\vspace*{-1em}
	\label{Table:compare-search-generally} 
\end{table*}

\vspace{-1em}
\subsection{Comparison between Single-Reward and Hybrid-Reward}
All existing medical report generation methods utilizing reinforcement learning employ CIDEr as the reward, which was proven to be the best reward in image captioning, while this has not been demonstrated in our task. 
Table~\ref{Table:report-reward} shows the experimental results of our ReMRG-X model with a different reward on the basis of X-LAN. Assuming that all natural language evaluation metrics are equally important, we take the sum of them as the final evaluation metric, denoted Score. Taking each metric as the reward separately, we can find that it usually has the highest score when it is used as a reward. Intuitively, we believe that the mixture of them will achieve improved results on all the metrics, which is verified in Table~\ref{Table:report-reward}, where SWHR achieves higher overall scores compared with single-reward-based models. Furthermore, since different metrics occupy different roles, we are motivated to search for the most suitable weight for the hybrid rewards. As can be observed, the results of the different weighted hybrid reward (DWHR) further improve the performance on almost all the metrics and achieve the highest final scores. 
Examples in Figure~\ref{fig.sample} also prove that our HReMRG-X generates more accurate (i.e., has more correctly matched terms) than that generated by XLAN, which simply utilizes CIDEr as the reward. Therefore, we employ the different weighted hybrid reward in our research.

\begin{table*}
\caption{
Results of comparing the proposed m-linear attention with the state-of-the-art attention mechanisms. 
}
	\centering   
	\begin{tabular}{l|l|cccccccc}
		\hline  
		Dataset & Model & BLEU-1 & BLEU-2 & BLEU-3 & BLEU-4 &CIDEr  & METEOR & ROUGE-L   & Score  \\ \hline 
		\multirow{5}{*}{\scriptsize{\textbf{IU X-Ray}}} 
&HReMRG & 0.3491 & 0.2245 & 0.1551& 0.1139&0.4489&0.1511&0.2925&1.7351\\
&HReMRG-Att2in~\cite{rennie2017self}  & 0.4071&0.2755&0.1925 &0.1329 &0.3636 &0.1884 &0.3652  & 1.9252\\
& HReMRG-AdaAtt~\cite{lu2017knowing} &0.3534&0.2453&0.174&0.1213&0.3217&0.1658&0.3472&1.7287\\
& HReMRG-AdaAttMO~\cite{lu2017knowing}&0.3836 & 0.2479&0.1761&0.1289&0.3154&0.1683&0.3309&1.7511\\
&HReMRG-X~\cite{Pan2020}&\textbf{0.4399}& \textbf{0.3081}&\textbf{ 0.2135} & \textbf{0.1466} & 0.4378 & 0.1942 & 0.374 & 2.1141\\
& HReMRG-M & 0.4322 & 0.3034 & 0.2113 & 0.1462 &\textbf{ 0.4929} &\textbf{0.1945} & \textbf{ 0.3795} & \textbf{2.1599} \\
\midrule 
\multirow{5}{*}{\scriptsize{\textbf{MIMIC-CXR}}}  
& HReMRG& 0.3084 & 0.2131 & 0.1611 & 0.1252 & 0.3164 & 0.1461 & 0.3383 & 1.6086 \\
&HReMRG-Att2in2 & 0.3664&0.2666 &0.2045 & 0.1585& 0.3401& 0.1705&0.3875&1.8941 \\
& HReMRG-AdaAtt &0.4173 & 0.3004&  0.2287 & 0.1734 & 0.3101  &0.1834 & 0.3838 & 1.9972\\
& HReMRG-AdaAttMO&0.2886 & 0.1936 & 0.1441 & 0.1125 &  0.4008 & 0.1486 &  0.3319 &  1.6201\\
&HReMRG-X &\textbf{0.4849}& \textbf{0.3429} & 0.2548 & 0.1897 & 0.3423 & \textbf{0.2092} & 0.3742 & 2.1980\\
& HReMRG-M & 0.4821&	0.3428&	\textbf{0.2558} &\textbf{0.1920}&	\textbf{0.3529} &0.2081 &\textbf{0.3752} &\textbf{2.2089} \\
\hline
	\end{tabular}
	\vspace{-1.5em}
\label{Table:attention}
\end{table*}

\vspace{-0.5em}
\begin{figure*}[t]
	\vspace{-1em}
	\hspace{-1em}
	\subfigure[Different Initial Learning Rate]{
		\begin{minipage}[t]{0.33\linewidth}
			\centering
			\includegraphics[width=2.3in]{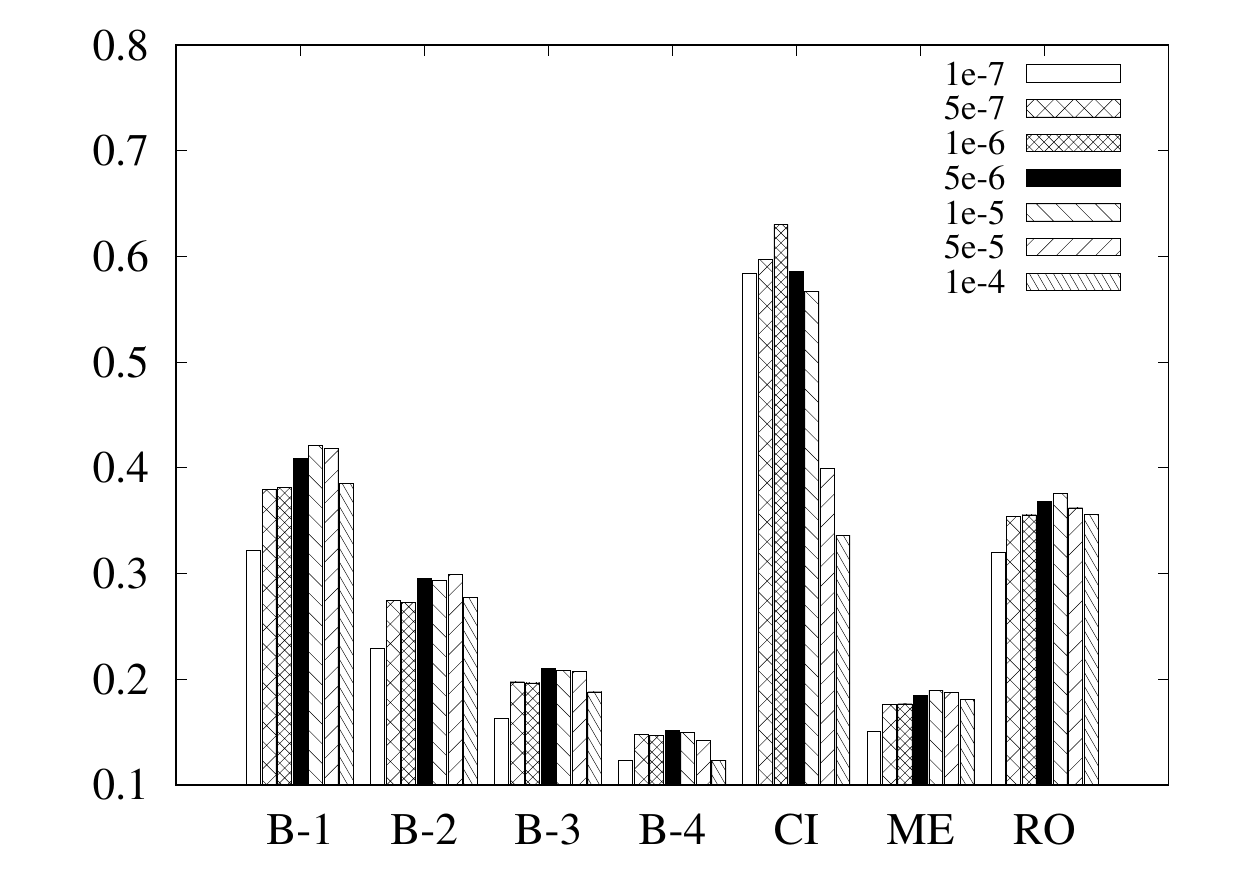}
		\end{minipage}%
	}%
	\subfigure[Different Learning Rate Decay Strategies]{
		\begin{minipage}[t]{0.33\linewidth}
			\centering
			\includegraphics[width=2.3in]{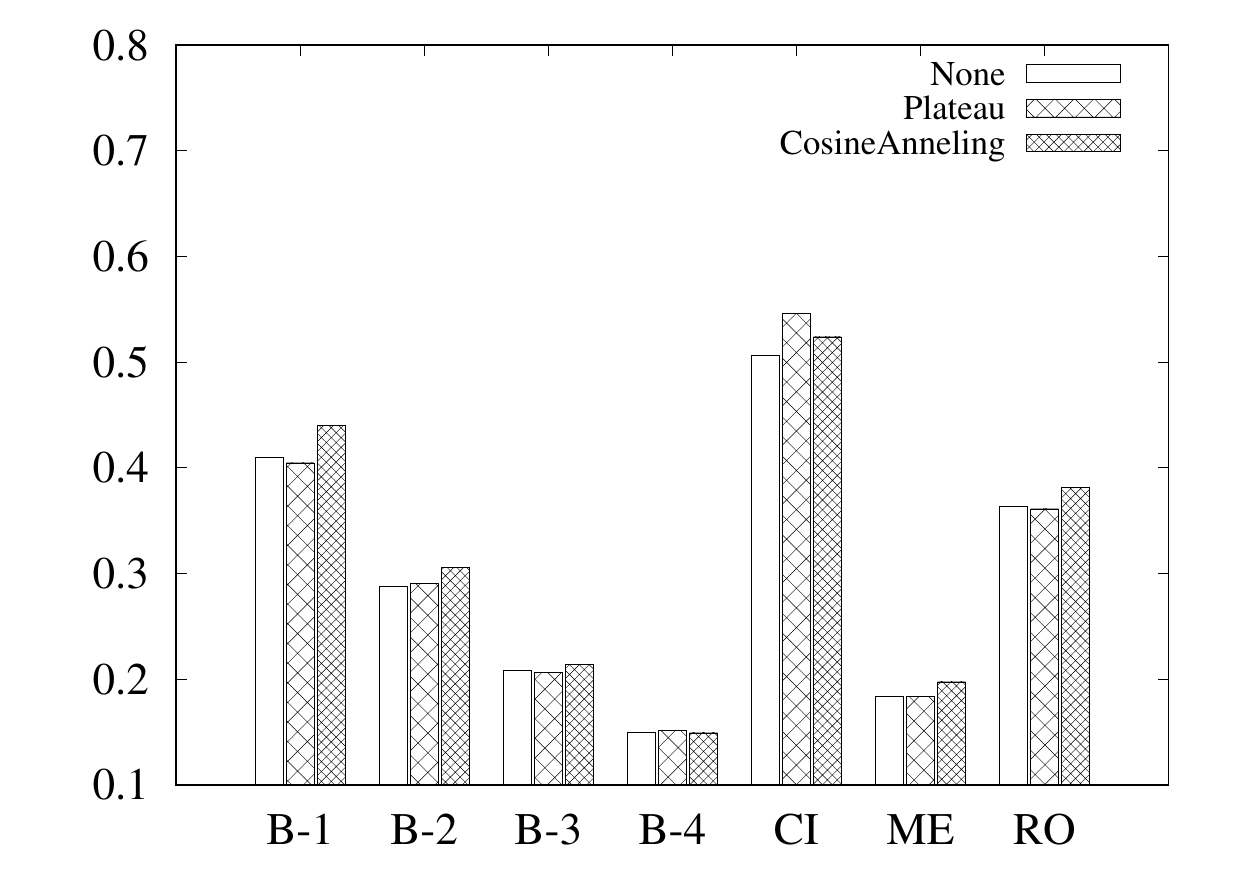}
		\end{minipage}%
	}%
	\subfigure[Different Optimizers]{
		\begin{minipage}[t]{0.33\linewidth}
			\centering
			\includegraphics[width=2.3in]{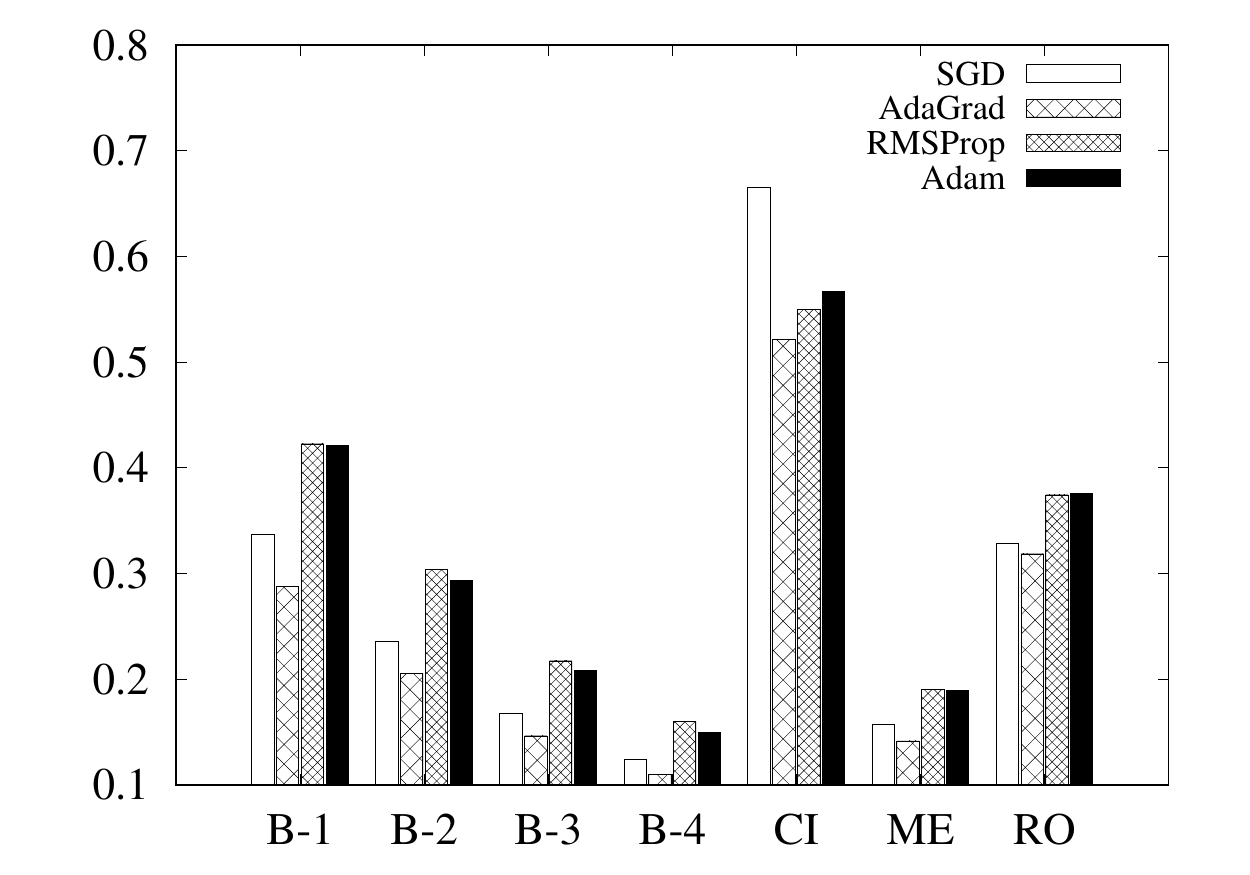}
		\end{minipage}%
	}%
	\vspace{-0.25em}
	\caption{Results of varying hyperparameters, where B, CI, ME, and RO denote BLEU, CIDEr, METEOR, and ROUGE-L, respectively.}
	\label{fig.parameters}\vspace*{-1ex}
	\vspace{-0.75em}
\end{figure*}

\vspace{-0.5em}
\subsection{Effectiveness of the Proposed Search Solution}

In order to demonstrate the effectiveness of our proposed search solution, we also compare the proposed method with generally adopted exponential search solutions. Since the complexity of seven rewards is too high, we select three of them for comparison, which are BLEU-4, CIDEr, and METEOR. Table~\ref{Table:compare-search-proposed} shows the results of our proposed search solution and Table~\ref{Table:compare-search-generally} shows the process of the original search. Comparing the two tables, we can find that the proposed method achieves the same optimal search results along with the original search, while our method largely saves the complexity from $O(n^3)$ to  $O(8)$, where n is 3. It proves that our proposed search solution can not only reduce the computation but can also achieve optimal results. The best result adopting three rewards is worse than that of searching seven rewards, which also demonstrates the necessity of adopting all the metrics.


\vspace{-1em}
\subsection{Effectiveness of Different Attention}

We use HReMRG as the baseline to further analyze the effectiveness of different attention.  According to Table~\ref{Table:attention}, we compare our proposed m-linear attention with four state-of-the-art attention mechanisms on the basis of the different weighted hybrid reward. As shown in Table~\ref{Table:attention}, with attention integrated, the performance of HReMRG-X boosts on almost all the metrics compared with the other three models, and our proposed HReMRG-M achieves the best performance according to the final score.
As Att2in2 is a simple attention mechanism integrating visual  and semantic information, AdaAtt employs an adaptive attention which decides when to attend to the visual positions or generate visually irrelevant words, and its improved version using maxout lstm on the basis of adaptive attention, all of them only consider single-dimension and low-order feature interactions. Differently from them, x-linear attention further employs a bi-linear pooling-based attention mechanism to achieve high-order feature reasoning and utilize channel-wise attention to further explore multi-dimensional feature interactions. Therefore, we conduct our research works on the basis of x-linear attention and propose m-linear attention, which employs a more suitable feature fusion method to further improve the overall performance.

\vspace{-1em}
\subsection{Influence of Different Hyperparameters}
We evaluate the influence of learning rate, strategies of learning rate, and optimizers. To obtain the optimal parameters of training, we train with HReMRG-MR model on IU X-ray for 100 epochs with a batch size of two.

To obtain the optimized initial learning rate, we compare the performance using different initial learning rates ranging from 1e-4 to 1e-7. Here, we employ no learning decay strategies and use the Adam optimizer. 
 According to Figure~\ref{fig.parameters}, the overall performance of 1e-5 and 5e-6 outperforms the others. Since the overall score of 1e-5 slightly outperforms that of 5e-6, we apply the initial learning rate of 1e-5 in our work.  
 
 Setting the initial learning rate to be 1e-5, we further investigate the influence of the strategy of learning rate decay when training with reinforcement learning by comparing the performances using \textit{Plateau} strategy,  \textit{CosineAnnealing} strategy, and that of not using learning rate decay strategies. Experimental results are presented in Figure~\ref{fig.parameters}.
 For \textit{Plateau} strategy, the learning rate is reduced by a factor of 0.8, once learning stagnates for 3 epochs. For the CosineAnnealing strategy, the cosine function is used as the learning rate annealing function, the maximum number of iteration epoch is set to be 10, and the minimum learning rate is $\rm{4e^{-8}}$. From Figure~\ref{fig.parameters}, we can observe that the performance of \textit{Plateau} is about the same as \textit{None}, and almost all evaluation scores using CosineAnnealing are beyond that using the Plateau decay strategy. Thus, we apply the CosineAnnealing learning rate decay strategy during reinforcement learning in our work.

Furthermore, we investigate the influence of optimizers by comparing SGD, AdaGrad, RMSProp, and Adam. Here, we set the initial learning rate to be 1e-5 and the learning rate decay strategy to be CosineAnneling. From Figure~\ref{fig.parameters}, 
We can observe that Adam and RMSProp perform best on the whole. Since the CIDEr score of Adam outperforms that of RMSProp a lot and CIDEr pays more attention to the important terms, we choose Adam in our experiments.

\vspace{-0.5em}
\section{Conclusion}
\label{sec5}

We have proposed to employ a different weighted hybrid reward to optimize the generated medical reports, we have also offered a search solution to obtain the best-weighted reward. Furthermore, we have also proposed to use m-linear attention,
which integrates bi-linear pooling to explore high-order feature interactions for intra-modal and inter-modal reasoning,
for medical report generation. Repetition penalty has also been utilized in reinforcement learning to generate long paragraphs.
We have conducted extensive experiments on two publicly
available datasets, IU X-Ray and MIMIC-CXR, which have
demonstrated the superior performance of our proposed approach to achieve state-of-the-art results.

Though our method achieved good results in overall optimization, its performance of generating more informative sentences could still be further improved.
As there are no effective metrics especially designed for
medical report generation, to evaluate our generated results
more effectively, we will explore a more reasonable and effective metric for medical report generation in future work.
While we have proposed a solution to search for the best hybrid reward and have reduced the search complexity greatly, it is still time-consuming to some extent.
In order to boost generation performance more effectively, inverse reinforcement learning will
also be taken into consideration to learn a more suitable reward for medical report generation in the future.



\ifCLASSOPTIONcaptionsoff
  \newpage
\fi



%


\vspace*{-0.5em}
\begin{small}
\bibliographystyle{IEEEtran}
\bibliography{report_abbr}

\begin{thebibliography}{10}
\providecommand{\url}[1]{#1}
\csname url@samestyle\endcsname
\providecommand{\newblock}{\relax}
\providecommand{\bibinfo}[2]{#2}
\providecommand{\BIBentrySTDinterwordspacing}{\spaceskip=0pt\relax}
\providecommand{\BIBentryALTinterwordstretchfactor}{4}
\providecommand{\BIBentryALTinterwordspacing}{\spaceskip=\fontdimen2\font plus
\BIBentryALTinterwordstretchfactor\fontdimen3\font minus
  \fontdimen4\font\relax}
\providecommand{\BIBforeignlanguage}[2]{{%
\expandafter\ifx\csname l@#1\endcsname\relax
\typeout{** WARNING: IEEEtran.bst: No hyphenation pattern has been}%
\typeout{** loaded for the language `#1'. Using the pattern for}%
\typeout{** the default language instead.}%
\else
\language=\csname l@#1\endcsname
\fi
#2}}
\providecommand{\BIBdecl}{\relax}
\BIBdecl

\bibitem{Wang_2018}
X.~Wang, Y.~Peng, L.~Lu, Z.~Lu, and R.~M. Summers, ``Tienet: Text-image
  embedding network for common thorax disease classification and reporting in
  chest {X}-rays,'' in \emph{CVPR}, 2018.

\bibitem{jing-etal-2018-automatic}
B.~Jing, P.~Xie, and E.~Xing, ``On the automatic generation of medical imaging
  reports,'' in \emph{ACL}, 2018, pp. 2577--2586.

\bibitem{xue2018multimodal}
Y.~Xue, T.~Xu, L.~R. Long, Z.~Xue, S.~Antani, G.~R. Thoma, and X.~Huang,
  ``Multimodal recurrent model with attention for automated radiology report
  generation,'' in \emph{MICCAI}, 2018, pp. 457--466.

\bibitem{10.1007/978-3-030-32692-0_77}
Y.~Xiong, B.~Du, and P.~Yan, ``Reinforced transformer for medical image
  captioning,'' in \emph{MLMI}, H.-I. Suk, M.~Liu, P.~Yan, and C.~Lian, Eds.,
  2019, pp. 673--680.

\bibitem{liu2019clinically}
G.~Liu, T.-M.~H. Hsu, M.~McDermott, W.~Boag, W.-H. Weng, P.~Szolovits, and
  M.~Ghassemi, ``Clinically accurate chest {X}-ray report generation,''
  \emph{arXiv preprint arXiv:1904.02633}, 2019.

\bibitem{li2018hybrid}
Y.~Li, X.~Liang, Z.~Hu, and E.~P. Xing, ``Hybrid retrieval-generation
  reinforced agent for medical image report generation,'' in \emph{NeurIPS},
  2018, pp. 1530--1540.

\bibitem{jing-etal-2019-show}
B.~Jing, Z.~Wang, and E.~Xing, ``Show, describe and conclude: On exploiting the
  structure information of chest {X}-ray reports,'' in \emph{ACL}, 2019, pp.
  6570--6580.

\bibitem{rennie2017self}
S.~J. Rennie, E.~Marcheret, Y.~Mroueh, J.~Ross, and V.~Goel, ``Self-critical
  sequence training for image captioning,'' in \emph{CVPR}, 2017, pp.
  7008--7024.

\bibitem{melas2018training}
L.~Melas-Kyriazi, A.~M. Rush, and G.~Han, ``Training for diversity in image
  paragraph captioning,'' in \emph{EMNLP}, 2018, pp. 757--761.

\bibitem{10.1007/978-3-030-26763-6_66}
L.~Sun, W.~Wang, J.~Li, and J.~Lin, ``Study on medical image report generation
  based on improved encoding-decoding method,'' in \emph{ICIC}, 2019, pp.
  686--696.

\bibitem{Chen_2020}
Z.~Chen, Y.~Song, T.-H. Chang, and X.~Wan, ``Generating radiology reports via
  memory-driven transformer,'' in \emph{EMNLP}, 2020, p. 1439–1449.

\bibitem{Pan2020}
Y.~Pan, T.~Yao, Y.~Li, and T.~Mei, ``X-linear attention networks for image
  captioning,'' in \emph{CVPR}, 2020, pp. 10\,971--10\,980.

\bibitem{kim2018bilinear}
J.-H. Kim, J.~Jun, and B.-T. Zhang, ``Bilinear attention networks,'' in
  \emph{NeurIPS}, 2018, pp. 1564--1574.

\bibitem{hu2018squeeze}
J.~Hu, L.~Shen, and G.~Sun, ``Squeeze-and-excitation networks,'' in
  \emph{CVPR}, 2018, pp. 7132--7141.

\bibitem{vedantam2015cider}
R.~Vedantam, C.~Lawrence~Zitnick, and D.~Parikh, ``{CIDEr: C}onsensus-based
  image description evaluation,'' in \emph{CVPR}, 2015, pp. 4566--4575.

\bibitem{8237362}
S.~Liu, Z.~Zhu, N.~Ye, S.~Guadarrama, and K.~Murphy, ``Improved image
  captioning via policy gradient optimization of {SPIDEr},'' in \emph{ICCV},
  2017, pp. 873--881.

\bibitem{anderson2018bottom}
P.~Anderson, X.~He, C.~Buehler, D.~Teney, M.~Johnson, S.~Gould, and L.~Zhang,
  ``Bottom-up and top-down attention for image captioning and visual question
  answering,'' in \emph{CVPR}, 2018, pp. 6077--6086.

\bibitem{demner2016preparing}
D.~Demner-Fushman, M.~D. Kohli, M.~B. Rosenman, S.~E. Shooshan, L.~Rodriguez,
  S.~Antani, G.~R. Thoma, and C.~J. McDonald, ``Preparing a collection of
  radiology examinations for distribution and retrieval,'' \emph{JAMIA}, pp.
  304--310, 2016.

\bibitem{johnson2019mimic}
A.~E. Johnson, T.~J. Pollard, S.~J. Berkowitz, N.~R. Greenbaum, M.~P. Lungren,
  C.-y. Deng, R.~G. Mark, and S.~Horng, ``{MIMIC-CXR}, a de-identified publicly
  available database of chest radiographs with free-text reports,''
  \emph{Scientific Data}, vol.~6, 2019.

\bibitem{papineni2002bleu}
K.~Papineni, S.~Roukos, T.~Ward, and W.-J. Zhu, ``{BLEU:} a method for
  automatic evaluation of machine translation,'' in \emph{ACL}, 2002, pp.
  311--318.

\bibitem{banerjee2005meteor}
S.~Banerjee and A.~Lavie, ``{METEOR: A}n automatic metric for {MT} evaluation
  with improved correlation with human judgments,'' in \emph{ACL}, 2005, pp.
  65--72.

\bibitem{lin2004rouge}
C.-Y. Lin, ``{ROUGE: A} package for automatic evaluation of summaries,'' in
  \emph{ACL}, 2004, pp. 74--81.

\bibitem{deng2009imagenet}
J.~Deng, W.~Dong, R.~Socher, L.-J. Li, K.~Li, and L.~Fei-Fei, ``{ImageNet: A}
  large-scale hierarchical image database,'' in \emph{CVPR}, pp. 248--255.

\bibitem{kingma2014adam}
D.~P. Kingma and J.~Ba, ``Adam: A method for stochastic optimization,''
  \emph{arXiv preprint arXiv:1412.6980}, 2014.

\bibitem{lu2017knowing}
J.~Lu, C.~Xiong, D.~Parikh, and R.~Socher, ``Knowing when to look: Adaptive
  attention via a visual sentinel for image captioning,'' in \emph{CVPR}, 2017,
  pp. 375--383.

\end{thebibliography}
\end{small}

\end{document}